\documentclass[10pt,twocolumn,letterpaper]{article}

\usepackage{iccv}
\usepackage{times}
\usepackage{epsfig}
\usepackage{graphicx}
\usepackage{amsmath}
\usepackage{amssymb}
\usepackage{url}
\usepackage{bm}
\usepackage{subfigure}
\usepackage{booktabs}
\usepackage{multirow}
\usepackage{bbding}
\usepackage{comment}
\usepackage{wrapfig}
\usepackage[table]{xcolor}

\usepackage[pagebackref=true,breaklinks=true,letterpaper=true,colorlinks,bookmarks=false]{hyperref}

\iccvfinalcopy 

\def\iccvPaperID{****} 

\ificcvfinal\fi

\begin{document}

\title{Oriented Feature Alignment for Fine-grained Object Recognition in High-Resolution Satellite Imagery}

\author{Qi Ming,  Junjie Song, Zhiqiang Zhou\\
 School of Automation, Beijing Institute of Technology\\
{\tt\small chaser.ming@gmail.com}
}

\maketitle

\ificcvfinal\thispagestyle{empty}\fi

\begin{abstract}
	Oriented object detection in remote sensing images has made great progress in recent years. However, most of the current methods only focus on detecting targets, and cannot distinguish fine-grained objects well in complex scenes. In this technical report, we analyzed the key issues of fine-grained object recognition, and use an oriented feature alignment network (OFA-Net) to achieve high-performance fine-grained oriented object recognition in optical remote sensing images. OFA-Net achieves accurate object localization through a rotated bounding boxes refinement module. On this basis, the boundary-constrained rotation feature alignment module is applied to achieve local feature extraction, which is beneficial to fine-grained object classification. The single model of our method achieved mAP of 46.51\% in the GaoFen competition and won 3rd place in the ISPRS benchmark with the mAP of 43.73\%. 
\end{abstract}


\section{Introduction}

With the increase of available remote sensing images, efficient interpretation of remote sensing images becomes more and more important. Object detection is an efficient interpretation method to process massive remote sensing images. In recent years, with the rapid development of deep learning, object detection technology has been developed by leaps and bounds \cite{ren2016faster,liu2016ssd,redmon2016you,girshick2014rich}. Object detection in remote sensing images has also made great progress. A series of high-efficiency detectors are proposed to achieve high-precision object detection in remote sensing images \cite{han2021align,ming2021cfc,ming2021sparse,ming2021optimization,yang2021rethinking,ding2019learning,ming2021dynamic,yang2021learning}.

Unlike general object detection that uses a horizontal bounding box (HBB) to annotate objects, object detection in remote sensing images often take oriented bounding boxes (OBB) to characterize objects. Compared with HBB, OBB contains less background, so it can more effectively describe the contour information of the targets, which is beneficial for the convolutional neural networks (CNNs) to distinguish boundary features. The current mainstream methods all preset rotated priori bounding boxes or generating rotated proposals, such as RoI-Transformer \cite{ding2019learning}, CFC-Net \cite{ming2021cfc}.

\begin{figure}[t]
	\subfigure[]{
		\includegraphics[width=0.22\textwidth]{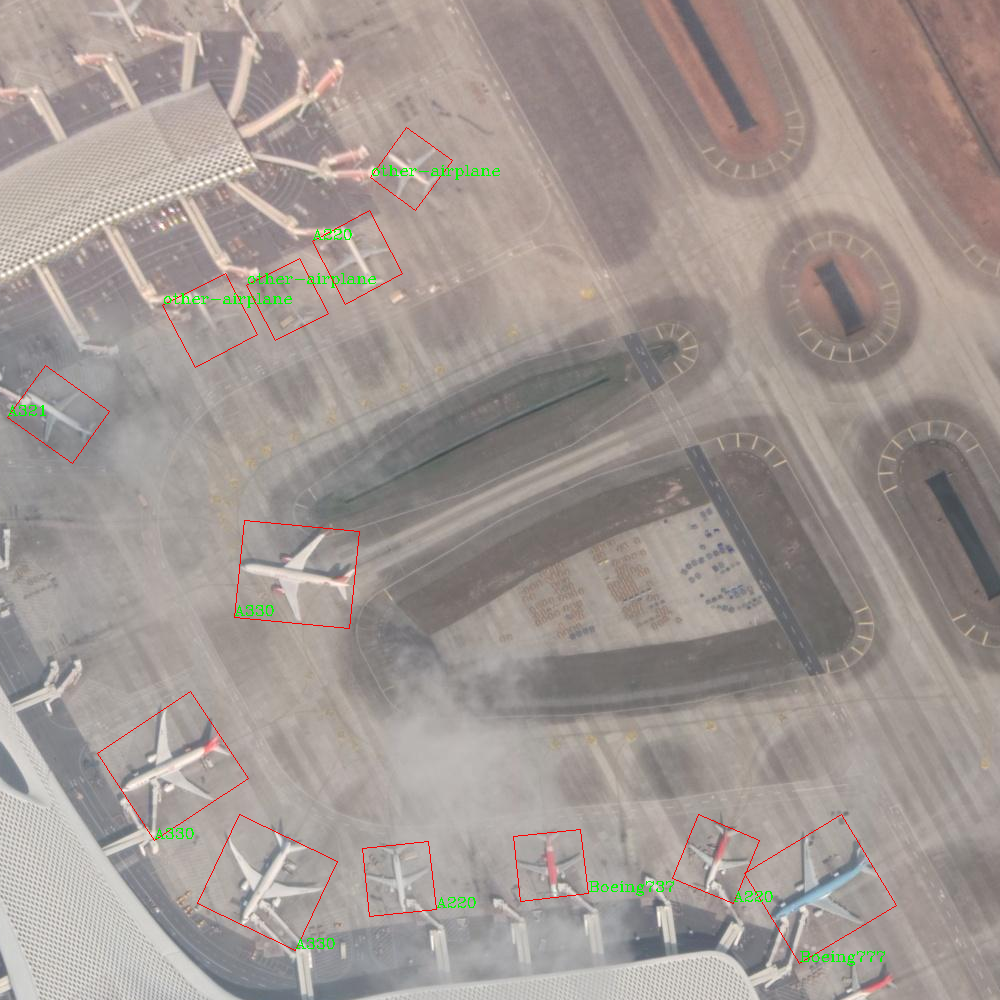}}\hspace{-1mm}
	\subfigure[]{
		\includegraphics[width=0.22\textwidth]{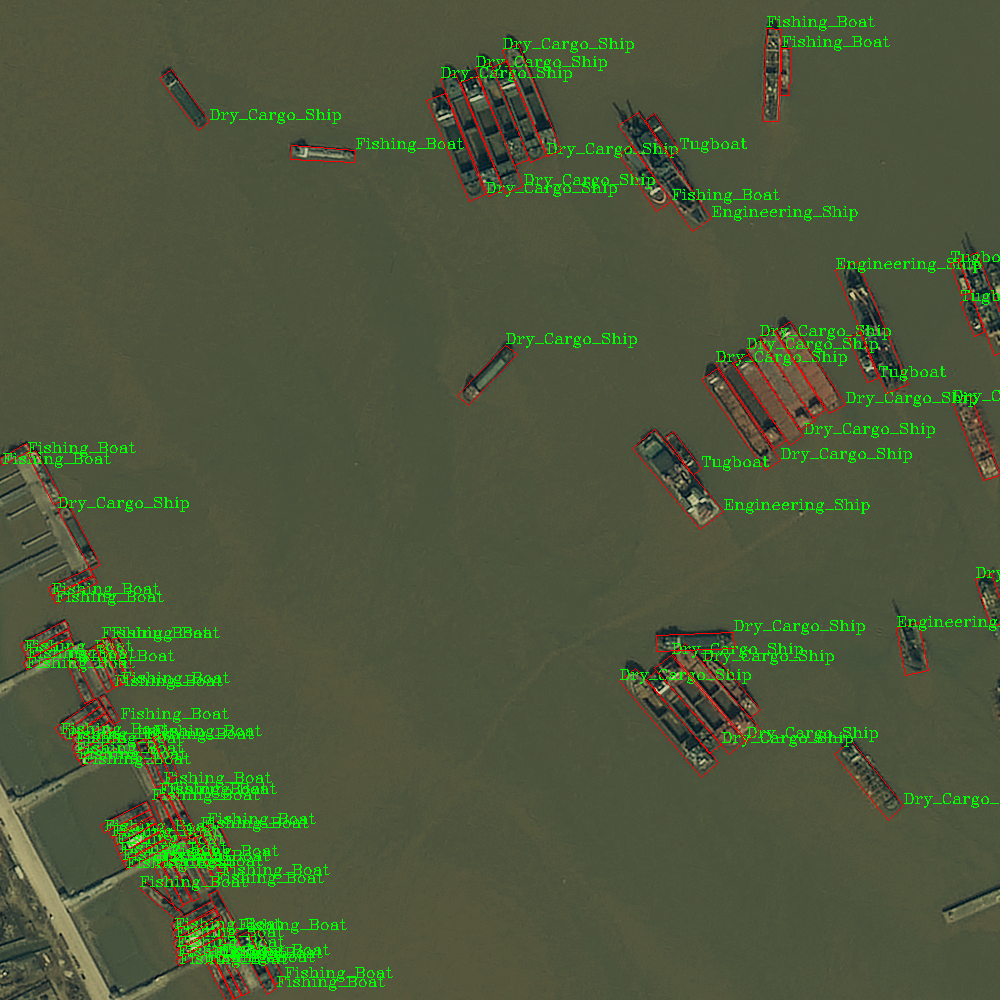}}
	\caption{Visualization of fine-grained aircraft recognition in optical remote sensing images.}
	\label{fig1}
\end{figure}

However, most of the current detectors inherit from the generic detection methods and do not pay attention to the unique problems in remote sensing images. That is, it is difficult to distinguish similar targets of different categories when viewed from a high altitude. For example, DOTA \cite{xia2018dota} is currently the largest remote sensing dataset with OBB annotations, which contains 15 categories. And there are big differences between the appearance of these categories. Even in remote sensing images, it is not hard to distinguish different categories (such as bridges and airplanes). These methods performed well on the DOTA dataset may achieve poor performance when distinguishing fine-grained targets. As shown in Figure \ref{fig1}, it is difficult to distinguish the different types of aircraft unless you are an expert in the field.

\begin{figure}[t]
	\begin{center}
	\includegraphics[width=1.0\linewidth]{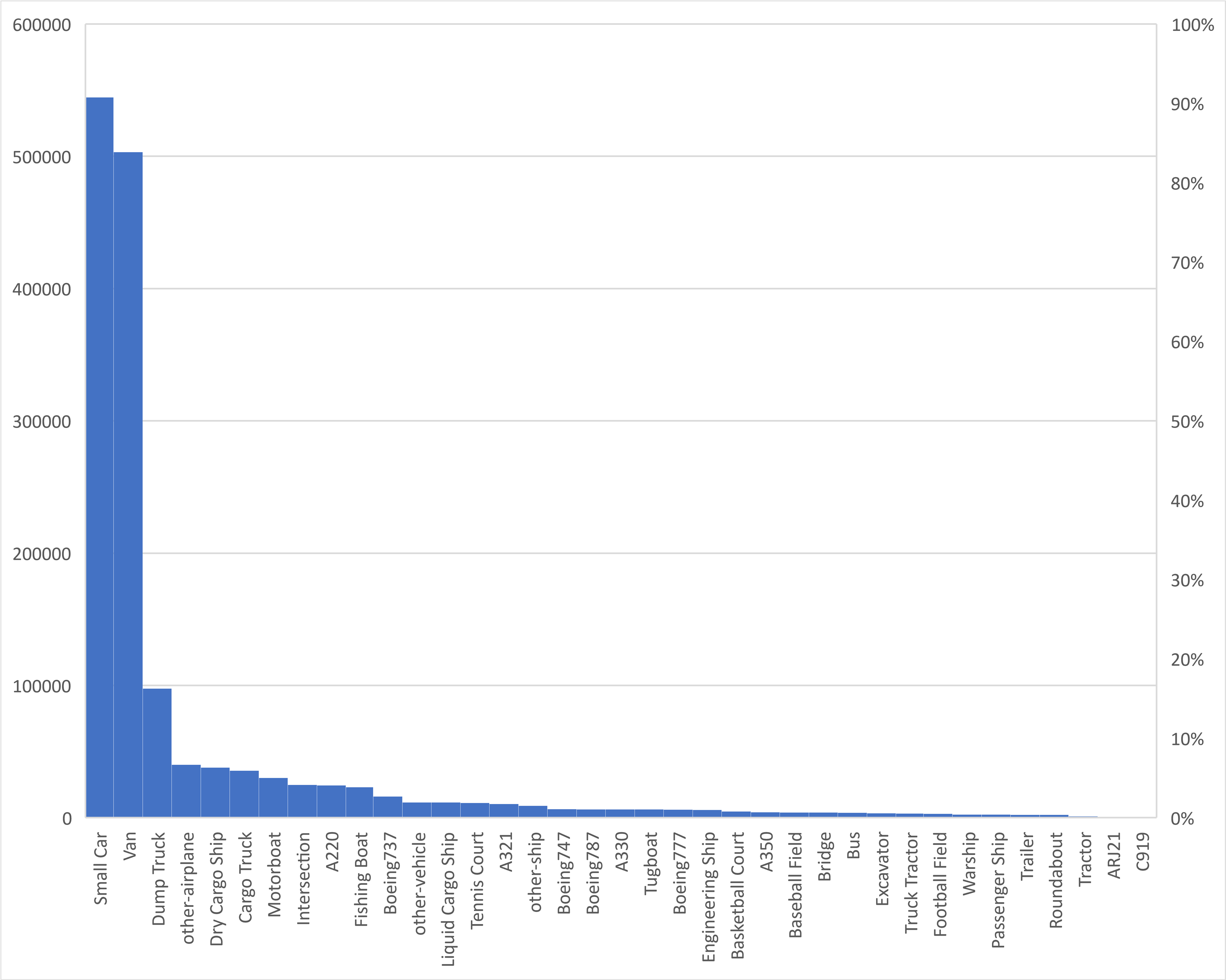}
	\end{center}
	   \caption{The distribution of the number of instances per category in the training set of FAIR1M.}
	\label{fig2}
	\end{figure}
	
Fine-grained object recognition in remote sensing images is a challenging task. Objects have few texture features at high viewing distances, and thus it is hard to identify them. Besides, the long-tail distribution of instances will further degrade the recognition accuracy. In this technical report, we use an oriented feature alignment network (OFA-Net) to achieve high-precision fine-grained object recognition in remote sensing images. OFA-Net is an anchor-free method that lays dense anchor boxes on the images. These anchor boxes generate candidate proposals to capture the possible positions of the objects to improve the recall rate. Specifically, we use an additional oriented box refinement module to adjust the position of the obtained proposals. For the classification task, we designed the oriented feature alignment branch to extract the features in the rotated proposal for fine-grained classification task.

Category imbalance is another problem that affects detection accuracy. As shown in Figure \ref{fig2}, there is a huge gap in the number of instances of different categories. In order to achieve a balanced training process, we performed class-balanced sampling to expand the dataset. At the same time, data augmentations\footnote{Implementation: \url{https://github.com/ming71/toolbox}} such as random cropping, flipping, optical distortion, etc. are used to expand the dataset. These methods can effectively improve the accuracy of recognition.

The main contributions of this technical report are as follows:

\begin{enumerate}
	\item  We analyzed the technical difficulties in fine-grained object recognition and developed solutions for it.
    \item OFA-Net is adopted to decouple the fine-grained object detection task into localization subtask and classification subtask. Then rotated anchor refinement module (RARM) and accurate detection module (ADM) are applied to achieves high-precision localization and classification.
    \item We designed an effective data augmentation strategy to greatly improve the recognition accuracy. Besides, we have tried series of tricks and reported the effects in detail.  

	\end{enumerate}

\section{Methodology}

\begin{figure}[t]
	\begin{center}
	\includegraphics[width=1.0\linewidth]{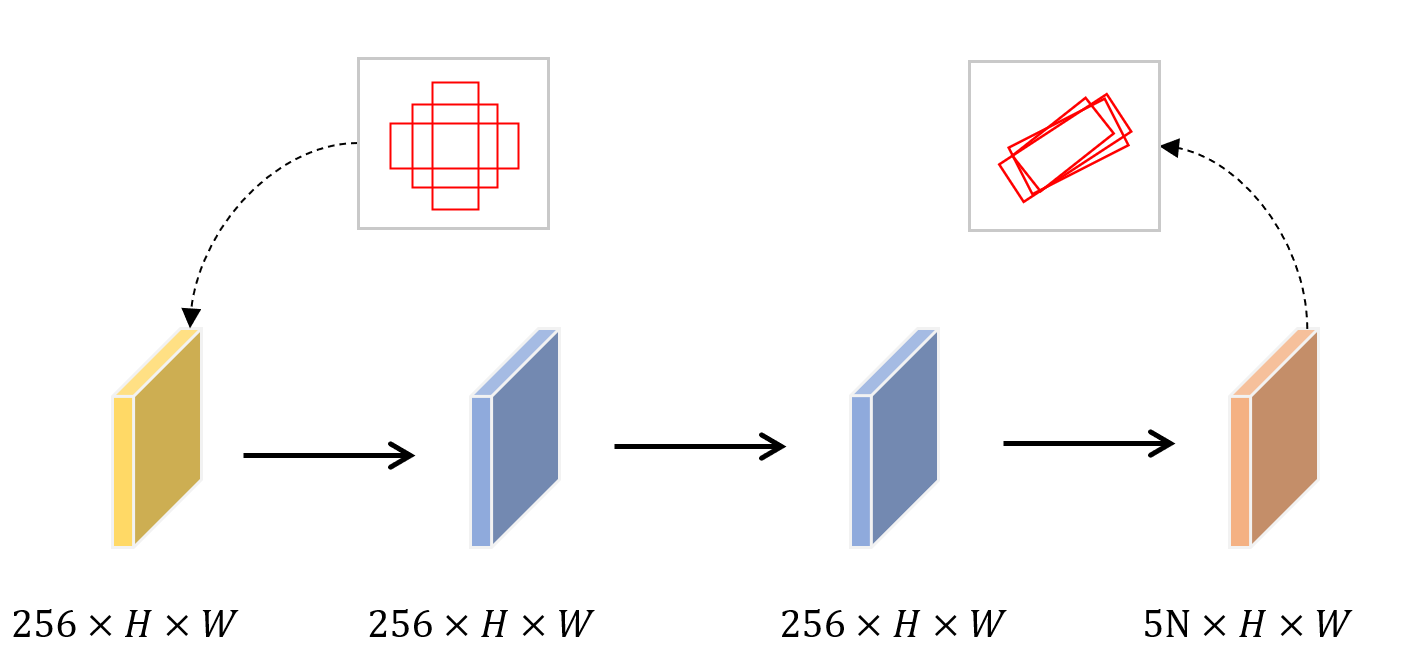}
	\end{center}
	   \caption{The structure of the proposal refinement module for bounding box localization branch.}
	\label{fig3}
	\end{figure}
We suggest that the fine-grained object recognition task can be decoupled into a localization task and a classification subtask. Then, we design  effective structures for the two subtasks to achieve high performance.

The purpose of the localization subtask is to achieve high recall and precision for bounding box regression. To this end, we adopt a cascade oriented refinement module in the detection branch. As shown in Figure \ref{fig3}, the localization branch consists of two parts: rotated anchor refinement module (RARM) and accurate detection module (ADM). Among them, RARM uses a low IoU threshold to select positive samples (set to 0.4 in our experiment), so as to improve the recall rate as much as possible. The high-quality proposals obtained after oriented anchor refinement are sent to the ADM for final localization output, thereby achieving high result accuracy. The superiority of the cascade refinement module has been confirmed in some previous work \cite{ming2021cfc,han2021align,yang2019r3det}.

\begin{figure}[t]
	\begin{center}
	\includegraphics[width=1.0\linewidth]{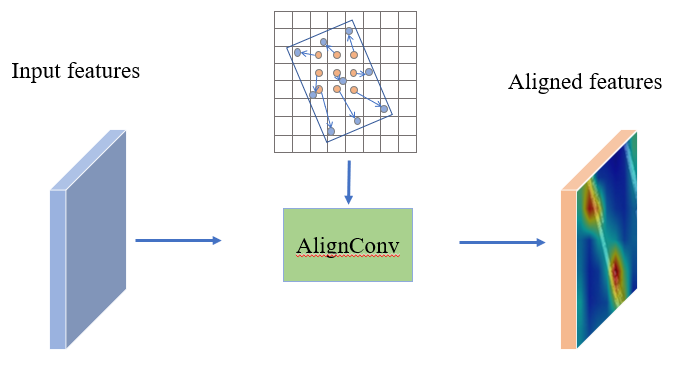}
	\end{center}
	   \caption{The detailed structure of AlignConv.}
	\label{fig4}
	\end{figure}

The classification task is one of the points of the fine-grained object recognition task. The fine-grained features need to be captured to achieve accurate discrimination for similar classes. Therefore, we suggest extracting as much effective texture information as possible, while ignoring the backgrounds. To this end, we use AlignConv in S$^2$ANet \cite{han2021align} to achieve the alignment and extraction of local features. AlignConv extracts aligned convolutional features from the proposal area that may contain objects, and extract texture information efficiently to help achieve high-performance classification. The details of AlignConv are shown in Figure \ref{fig4}.

\begin{table*}[t]
	\caption{ Performance evaluation on FAIR1M dataset.}
	\small
		\centering
		\setlength{\tabcolsep}{1.5mm}{
	\begin{tabular}{c|ccccccccccc}
			\toprule
			ID &  1 & 2 & 3 & 4 & 5 & 6 & 7 & 8 & 9 & 10 & 11  \\ 
			\midrule
			Epoch  & 12  & 6 & 6 & 12 & 6 & 12 & 12 & 9 & 6 & 6 & 6 \\
			Backbone & R-50 & R-50 & R-50 & R-50 & R-50 & R-50 & R-50 & R-50 & R-50 & R-50 & R-50 \\
			IoU thres & 0.5/0.5 &  0.5/0.5 &  0.5/0.5 &  0.5/0.5 &  0.5/0.5 &  0.5/0.5 &  0.5/0.5 &  0.5/0.5 &  0.5/0.6 &  0.5/0.6 &  0.5/0.6 \\
			Lr  &  0.01  &  0.01  &  0.01  &  0.01  &  0.01  &  0.01  &  0.01  &  0.01  &  0.01  &  0.02  &  0.01  \\
			\midrule
			DOTA pretrain &  &  &  &  & $\checkmark$ & $\checkmark$ & $\checkmark$ & $\checkmark$ & $\checkmark$ & $\checkmark$ & $\checkmark$ \\
			GF AIR pretrain &  &  &  &  & $\checkmark$ &  & $\checkmark$ & $\checkmark$ & $\checkmark$ & $\checkmark$ & $\checkmark$ \\
			Augmentation &  &  &  &  &   &  &  &  &  &  &  $\checkmark$ \\
			\midrule
			mAP  & 40.7887	&39.4929 &	39.0654  &	40.9713  &	40.6944 &	40.7576  &	41.5011  &	41.5650  &	41.3631  &	40.8882  &	43.7304 \\
			\midrule
			B737   &39.7084	 &37.9241 &	37.5311  &	40.0455  &	40.4168 &	37.4206  &	41.0263  &	41.3432 	 &41.3122  &	39.8741  &	43.2032 \\
			B747& 84.8311&	84.672&	84.2292 &	84.8716 &	84.7938	&84.5244 	&85.1451& 	85.4570 	&84.2687 	&84.6369 	&87.5791 \\
			B777& 16.0072	&13.4221&	12.3952 &	16.2657 &	20.777	&15.8530 &	20.1906 	&19.4324 	&25.4138 &	17.4337 &	23.5836 \\
			C919 &27.6657&	29.6156&	29.4787 &	27.7799 &	0.7919&	22.4250 &	26.5894 &	28.6857 &	23.2224 &	20.6797 	&30.7125 \\
			A220 & 51.3761&	49.8609&49.6541 &	52.3854& 	52.5156	&52.4346 &	52.9216 &	53.3567 	&52.3169 	&50.0491 	&54.3105 \\
			A321 & 68.3039 &	62.8309 &	62.1537 & 	68.8290  &	70.3504	 &66.3956  &	68.7606 	 &68.8555  &	69.2549  &	67.6381 	 &67.8270 \\
			A330&	72.6854	&65.7713&	65.5798 &	72.7667 &	75.8461	&66.5879 &	73.7539 	&73.1206 	&73.0727 &	72.1643 &	73.6505 \\
			A350	&70.3734	&66.65	&65.9373 &	70.3735 &	75.6009&	71.2745 &	74.2873 &	73.8647 	&73.7684 &	73.2839 &	73.5631 \\
			ARJ21	&26.2759&	30.3777&	29.8765 &	26.4851 &	26.3329&	30.6652 &	28.8742 	&28.8546 	&27.5266 &	28.8618 &	29.2502 \\
			PS&	12.1353	&11.1162&	10.7433 &	12.2696 &	9.9621&	11.0566 	&9.1320 &	9.3924 &	10.0265 &	8.2911 &	9.8989 \\
			MB&	58.2737&	58.1658	&57.6889 &	58.6086 &	60.9869	&61.2956 &	61.4033 &	61.5985 	&61.0369 &	62.3887 &	68.1542 \\
			FB&	8.8339&	6.9411&	6.6453 &	8.9604& 	9.56&	9.5880 &	9.9439& 	10.1361 &	9.7975 &	10.2011 &	11.6979 \\
			TB	&27.0258	&28.8266&	28.5602 	&27.1061 &	20.2742&	22.4229 	&23.6130 &	24.2323 &	25.9330 &	25.9567 	&30.3681 \\
			ES &	9.2314 &	10.562 &	10.2186  &	9.3889 	 &11.0026	 &11.7107  &	9.6942  &	9.9039  &	9.5188  &	8.0573  &	11.8250 \\
			LCS	 &21.8006 &	21.4709	 &20.8064  &	21.9911  &	23.2248 &	20.8399  &	22.5316 	 &21.8622 	 &21.3641  &	21.7128  &	26.8041 \\
			DCS	 &39.3406	 &35.155	 &34.4924  &	39.6285 	 &38.8381	 &38.5661  &	38.2908 	 &38.3285 	 &41.1210  &	41.1528  &	38.6832 \\
			WS &	30.5194	 &21.9942	 &19.8088 	 &31.4485 	 &34.4608	 &34.8139  &	36.5231  &	35.8599  &	36.6920  &	35.9102 	 &35.4155 \\
			SC	 &67.3107 &	68.9919	 &68.8422 	 &67.3881  &	65.2224 &	66.1761 	 &65.3095  &	65.5542  &	65.5790 	 &65.7107 	 &69.6212 \\
			Bus	&47.237&	44.5228	&44.1516&47.2995&43.4714&	37.5358&36.1380&37.1110&38.9024&43.3436&33.6659 \\
			CT&	43.2981	&44.4982&	43.4257&43.4072&43.9681&	43.2695&43.1941&43.2861&43.8268&43.5153&44.6937 \\
			DT&	45.3287&	46.1024&	45.8578&45.4140&40.7919	&43.9990&41.5088&41.7621&40.3844&40.4241&49.2335 \\
			Van	&65.2316&	66.304&	66.1225&65.2860&62.7584	&64.4427&62.6035&62.9158&62.5761&62.6648&71.6140 \\
			Trailer	&11.6865&	13.081&	12.8507&11.7254&12.7795&	14.6236&14.9420&14.8581&11.7114&12.8089&13.4503 \\
			Tractor	&1.6189	&1.4166	&1.2066&1.6343&3.2018	&2.2870&2.9026&2.8953&1.6872&2.3507&6.0550 \\
			Excavator	&17.28&	18.2321	&17.8421&17.3518&14.7494	&15.7700&16.0721&17.1267&14.4945&12.3239&17.3645 \\
			TT	&5.2196	&3.2102	&3.0169&5.2491&2.9084	&5.3504&2.2923&1.7956&1.0404&1.8090&5.3458 \\
			BC	&38.6951&	35.4245&	34.9380&38.9421&42.3296	&45.9467&42.1421&41.3522&41.5195&41.1315&49.2033 \\
			TC	&76.8179&	76.197&	76.0087&76.9637&87.4435	&80.8257&87.2986&87.0596&86.7541&87.1786&83.7470 \\
			FF&	62.8118	&61.597&	61.2147&62.9605&65.1404&	62.9778&64.9985&65.0109&63.4872&62.5057&61.2931 \\
			BF&	87.4028	&84.3917	&84.0975&87.5050&87.6035&	86.5839&87.7207&87.8594&87.7355&87.8116&88.7665 \\
			Intersection&	55.9009	&55.1087&	54.8128&55.9881&54.2318	&57.1986&54.9236&54.4243&53.5892&53.1332&57.9687 \\
			Roundabout&	19.5186&	20.2041	&20.1930&19.5463&13.763	&20.7725&18.6329&18.4380&20.1613&18.9133&24.6793 \\
			Bridge&	24.7308	&24.1289	&24.0228&24.7419&33.3394&	31.2905&33.2344&33.8300&37.2813&35.8941&35.8224 \\
			\bottomrule
		\end{tabular}}
	\label{table1}
	\end{table*}

Moreover, in order to further improve the performance of fine-grained object recognition, we have tried many tricks. To solve the problem of long-tail distribution in fine-grained object recognition, we adopted class-balanced sampling on the original dataset. It’s known that some categories such as ARJ21 have relatively fewer instances, we resample them to expand the dataset. Secondly, the scale of some objects (such as ships) varies greatly in different scenarios. We used multi-scale training and testing to make the network adapt to variations in different scales. Finally, to enable the network to learn a certain degree of translation and rotation invariance, we adopted the data augmentations of affine transformations.

There are more data analysis and processing techniques such as pre-training weights, training schedule, and adjustment of the learning rate, etc. We will give them in detail in the experimental part.

\section{Experimental Results}

\subsection{Dataset and Implementation Details}
The data set we used includes the fine-grained object recognition competition in the GaoFen Challenge and FAIR1M dataset \cite{sun2021fair1m}. FAIR1M is the ISPRS benchmark on object detection in high-resolution satellite images which contains 37 classes. The object categories in the FAIR1M dataset are Boeing 737, Boeing 777, Boeing 747, Boeing 787, Airbus A320, Airbus A220, Airbus A330, Airbus A350, COMAC C919, COMAC ARJ21, other-airplane, passenger ship, motorboat, fishing boat, tugboat, engineering ship, liquid cargo ship, dry cargo ship, warship, other-ship, small car, bus, cargo truck, dump truck, van, trailer, tractor, truck tractor, excavator, other-vehicle, baseball field, basketball court, football field, tennis court, roundabout, intersection, and bridge. In the dataset, each object is annotated by an oriented bounding box (OBB). We conducted evaluation on the test set of the FAIR1M dataset and submit the final submission on servers for GaoFen Challenge.

The input images are cropped into 800*800 patches with a gap of 150. We use the SGD optimizer to train the network with a learning rate of 0.05. We train the models for 12 epochs on 2 RTX 2080ti GPUs. Random cropping, flipping, affine transformation, and optical distortion are used for data augmentation. We train the model on the train set of FAIR1M and eval on the test set on FAIR1M. Finally, we test the model on the GaoFen challenge. The total results on FAIR1M are shown in Table \ref{table1}.

\subsection{Evaluation of pretrain models}
We have tried to use additional data for pre-training, including the DOTA dataset and 300 images of fine-grained aircraft recognition data in the GaoFen2020 competition, and then fine-tune the model on FAIR1M. Experiments of ID1 and ID6 in Table \ref{table1} (40.7887\% vs. 40.7576\%) show that the DOTA pre-training weight provides good prior knowledge in some categories with higher scene similarity (such as baseball field, basketball court, football field, tennis court). The performance of the related classes is improved. On the other hand, fine-grained aircraft images are very similar to FAIR1M data. Experiments of ID5 and ID6 in Table \ref{table1} show that the use of a pre-trained model effectively improves the performance of aircraft recognition. Moreover, the pre-trained model speeds up the model convergence. For example, with the DOTA pre-training weights, the model fine-tuned for 1 epoch on FAIR1M can achieve an accuracy comparable to 6 epoch training from scratch.

\subsection{Evaluation of IoU threshold for training sample selection}
Different IoU thresholds will lead to various training sample distributions and affect the detection performance. A low IoU threshold for positive samples helps to improve the recall rate but reduces the detection accuracy. A high IoU threshold will achieve higher accuracy, but the recall rate may not be good. We have tried different threshold settings. As shown in ID5 and ID9 in Table \ref{table1} (40.6944\% vs. 41.3631\%), we can a low IoU threshold in the first stage to ensure the recall rate, and set a high IoU threshold in the second stage to improve accuracy can achieve higher detection performance.

\subsection{Evaluation of detection confidence}
\label{sec}
The model using a high detection confidence threshold can output more credible detection results, while the one using a low detection threshold achieves a higher recall rate. We need to make trade-offs among them. The experimental results of ID1 (40.7887\%) compared with ID4 (40.9713\%) and ID2 (39.4929\%) compared with ID3 (39.0654\%) show that a lower confidence threshold can get higher mAP. 

\subsection{Evaluation of data augmentation}

We use a variety of data augmentation methods, including random flips, affine transformations, and optical distortions. Methods such as cutout, random noise, and random pixel dropping have also been tried, but they did not work well. In addition, we resample categories with few instances to expand the training set. With these augmentations, we doubled the dataset. Then multi-scale training and testing were performed, and a substantial performance improvement was achieved, as shown by ID10 and ID11 (40.8882\% vs. 43.7304\%). This single model achieved the mAP of 46.1747\% in the GaoFen fine-grained aircraft recognition track (ranking 10/213), improved by more than 6 points compared with the baseline.

\subsection{Outlook and other analysis}

The above experimental results are obtained from the single-model evaluation. Due to constraints of time and available GPU resources, we are not able to further apply complex data augmentation and multi-model ensemble for higher mAP. And we believe our methods and training strategies could have achieved better performance.

We have also tried other strategies. For example, the single model ensemble helps to slightly improve performance (about 0.5 points in the GaoFen challenge). It's also interesting that the expert model for certain categories cannot significantly improve the detection accuracy even complex data augmentations are attached to the tiny data. Maybe caused by overfitting? or the misjudgment between similar classes? I'm not sure. The first thing is to figure out whether the classification or detection is a real problem, or the both. I'll make it clear in the future. There's one more thing, class-agnostic NMS never works. Obviously, recall is more important than precision in mAP in most cases, which is consistent with the conclusion in Section \ref{sec}.

\section{Conclusion}
In this technical report, we analyzed the difficulties of fine-grained object recognition in optical remote sensing images and designed effective strategies to achieve high-precision object detection. Specifically, we decompose the fine-grained object recognition detection task into the detection subtask and the classification subtask. The rotated anchor refining module is designed to obtain accurate object positioning, while the oriented feature alignment module is applied to effectively extract the texture features of the object. Besides, we designed customized data augmentation strategies and resampling strategies to alleviate the problem of category imbalance. Even the single model of our methods achieved impressive results. There is still a lot of room for improvement and we will figure out the solutions in the future.


{
\bibliographystyle{ieee_fullname}
\bibliography{egbib}
}
\end{document}